\documentclass[a4paper, oneside, twocolumn, notitlepage, 10pt]{extarticle_ecoc}
\usepackage{ecoc}
\usepackage[printonlyused,nolist]{acronym}
\usepackage{bm}
\usepackage{tikz}
\usepackage{pgfplots}
\pgfplotsset{compat=1.18}
\definecolor{kit-green100}{rgb}{0,.59,.51}
\definecolor{kit-green70}{rgb}{.3,.71,.65}
\definecolor{kit-green50}{rgb}{.50,.79,.75}
\definecolor{kit-green30}{rgb}{.69,.87,.85}
\definecolor{kit-green15}{rgb}{.85,.93,.93}
\definecolor{KITgreen}{rgb}{0,.59,.51}

\definecolor{KITpalegreen}{RGB}{130,190,60}
\colorlet{kit-maigreen100}{KITpalegreen}
\colorlet{kit-maigreen70}{KITpalegreen!70}
\colorlet{kit-maigreen50}{KITpalegreen!50}
\colorlet{kit-maigreen30}{KITpalegreen!30}
\colorlet{kit-maigreen15}{KITpalegreen!15}

\definecolor{KITblue}{rgb}{.27,.39,.66}
\definecolor{kit-blue100}{rgb}{.27,.39,.67}
\definecolor{kit-blue70}{rgb}{.49,.57,.76}
\definecolor{kit-blue50}{rgb}{.64,.69,.83}
\definecolor{kit-blue30}{rgb}{.78,.82,.9}
\definecolor{kit-blue15}{rgb}{.89,.91,.95}

\definecolor{KITyellow}{rgb}{.98,.89,0}
\definecolor{kit-yellow100}{cmyk}{0,.05,1,0}
\definecolor{kit-yellow70}{cmyk}{0,.035,.7,0}
\definecolor{kit-yellow50}{cmyk}{0,.025,.5,0}
\definecolor{kit-yellow30}{cmyk}{0,.015,.3,0}
\definecolor{kit-yellow15}{cmyk}{0,.0075,.15,0}

\definecolor{KITorange}{rgb}{.87,.60,.10}
\definecolor{kit-orange100}{cmyk}{0,.45,1,0}
\definecolor{kit-orange70}{cmyk}{0,.315,.7,0}
\definecolor{kit-orange50}{cmyk}{0,.225,.5,0}
\definecolor{kit-orange30}{cmyk}{0,.135,.3,0}
\definecolor{kit-orange15}{cmyk}{0,.0675,.15,0}

\definecolor{KITred}{rgb}{.63,.13,.13}
\definecolor{kit-red100}{cmyk}{.25,1,1,0}
\definecolor{kit-red70}{cmyk}{.175,.7,.7,0}
\definecolor{kit-red50}{cmyk}{.125,.5,.5,0}
\definecolor{kit-red30}{cmyk}{.075,.3,.3,0}
\definecolor{kit-red15}{cmyk}{.0375,.15,.15,0}

\definecolor{KITpurple}{RGB}{160,0,120}
\colorlet{kit-purple100}{KITpurple}
\colorlet{kit-purple70}{KITpurple!70}
\colorlet{kit-purple50}{KITpurple!50}
\colorlet{kit-purple30}{KITpurple!30}
\colorlet{kit-purple15}{KITpurple!15}

\definecolor{KITcyanblue}{RGB}{80,170,230}
\colorlet{kit-cyanblue100}{KITcyanblue}
\colorlet{kit-cyanblue70}{KITcyanblue!70}
\colorlet{kit-cyanblue50}{KITcyanblue!50}
\colorlet{kit-cyanblue30}{KITcyanblue!30}
\colorlet{kit-cyanblue15}{KITcyanblue!15}

\usepackage{comment}
\usepackage{pgfplots}
\usepackage{subcaption}
\usepackage{url}

\begin{document}
\selectlanguage{english}    %
\begin{acronym}
    \acro{ann}[ANN]{artificial neural network}
    \acro{ber}[BER]{bit error rate}
    \acro{bptt}[BPTT]{backpropagation through time}
    \acro{cd}[CD]{chromatic dispersion}
    \acro{ce}[CE]{cross-entropy}
    \acro{dsp}[DSP]{digital signal processing}
    \acro{lif}[LIF]{leaky-integrate-and-fire}
    \acro{li}[LI]{leaky-integrate}
    \acro{imdd}[IM/DD]{intensity modulation/direct detection}
    \acro{mac}[MAC]{multiply–accumulate}
    \acro{ml}[ML]{machine learning}
    \acro{pgt}[PGT]{policy gradient theorem}
    \acro{pg}[PG]{policy gradient}
    \acro{pam}[PAM]{pulse amplitude modulation}
    \acro{sgd}[SGD]{stochastic gradient descent}
    \acro{snn}[SNN]{spiking neural network}
    \acro{te}[TE]{ternary encoding}
    \acro{ttfs}[TTFS]{time-to-first-spike}
    \acro{rl}[RL]{reinforcement learning}
\end{acronym}

\title{Encoding Optimization for Low-Complexity Spiking Neural Network Equalizers in IM/DD Systems}%

\author{
    Eike-Manuel Edelmann, Alexander von Bank,
    Laurent Schmalen
}

\maketitle                  %

\begin{strip}
    \begin{author_descr}

        Communications Engineering Lab (CEL), Karlsruhe Institute of Technology (KIT), 
        \textcolor{blue}{\uline{edelmann@kit.edu}} 

    \end{author_descr}
\end{strip}

\renewcommand\footnotemark{}
\renewcommand\footnoterule{}

\renewcommand{\thefootnote}{}

\begin{strip}
    \begin{ecoc_abstract}
        Neural encoding parameters for spiking neural networks (SNNs) are typically set heuristically. We propose a reinforcement learning-based algorithm to optimize them. Applied to an SNN-based equalizer and demapper in an IM/DD system, the method improves performance while reducing computational load and network size.
    \end{ecoc_abstract}
\end{strip}

\section{Introduction}
To keep up with the rapid growth of data center traffic, optical transceivers must simultaneously deliver higher data rates and energy-efficiency~\cite{böcherer}.
Using \ac{ml} algorithms, higher data rates can be achieved~\cite{Schmalen_advances}. 
To realize energy-efficient transceivers, recent research explores shifting portions of \ac{dsp} to more power-efficient frontends, e.g., neuromorphic hardware~\cite{böcherer, Brainscale2}.
Emulated on neuromorphic hardware, \acp{snn} promise to combine the performance of \ac{ml} with the energy efficiency of neuromorphic computing~\cite{Neftci_lessons}.
Recent research highlights the potential of \ac{snn}-based \ac{dsp} for optical communication systems~\cite{arnold_sppcom,arnold_journal,böcherer,vonBank_sppcom}. 
For an \ac{imdd} link affected by non-linear impairments and chromatic dispersion, an \ac{snn}-based equalizer and demapper outperforms both linear and traditional \ac{ann}-based equalizers~\cite{arnold_sppcom}. 
In~\cite{arnold_journal}, the performance of the equalizer was replicated using neuromorphic hardware.
In addition,~\cite{böcherer} applied the equalizer to experimental \ac{imdd} data.

\acp{snn} process information through sparse event-driven spikes, interchanged between their neurons.
Neural encoding must be applied to convert real-world data into sequences of spikes.
Multiple encoding approaches exist~\cite{auge}, each having various non-differentiable parameters that are typically determined heuristically.
While \cite{arnold_sppcom,arnold_journal,böcherer,vonBank_sppcom} optimize the \ac{snn}, the parameters of the neural encoding are determined heuristically, possibly selecting bad parameters.
In~\cite{vonBank_ofc}, a neural encoding that outputs temporal sequences of real numbers is successfully optimized.
However, the binary spiking character of the output is lost.

In this paper, we modify the \ac{rl}-based method of~\cite{aoudia18} to optimize the parameters of the neural encoding.
Using the neural encoding of~\cite{arnold_journal}, we show that our approach yields an efficient encoding, which reduces the computational load and the size of the \ac{snn} while maintaining the same performance.
Compared to Bayesian optimization, our approach yields improved performance, and avoids the time-consuming optimization of various \acp{snn} with different sets of encoding parameters.
The code is available at~\cite{code}.

\section{Spatio-temporal Neural Encoding}
\newcommand{\Kmax}{K}
Let $x\in\mathbb{R}$ denote an arbitrary number.
To make $x$ interpretable by the \ac{snn}, the neural encoding converts $x$ into $J$ parallel spike signals ${z_j(t)\in\{0,1\},\, j=1,2,\ldots,J}$.
Each spike signal is fed to an input neuron of the \ac{snn}.
Thus, the encoding has a spatial dimension $j$ and a temporal dimension $t$. 
All spike signals ${z_j(t)}$ can be generated independently, using, e.g., temporal delay methods, which encode the information in the relative timing between the first spike of the signal and a reference~\cite{auge}.

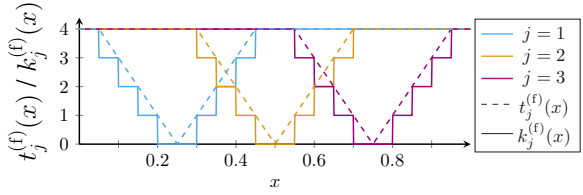
\begin{figure}
    \centering
    \resizebox{.48\textwidth}{!}{\begin{tikzpicture}[very thick,>=latex]
  \begin{axis}[
    xlabel={$x$},
    ylabel={\Large $t_j^{(\mathrm{f})}(x)$ / $k_j^{(\mathrm{f})}(x)$},
    xmin=0, xmax=1,
    ymin=0, ymax=4.4,
    domain=0:1,
    samples=200,
    axis lines=left,
    legend style={at={(1.01,.99)}, anchor=north west, legend columns=1,opacity=.9},
    width=9cm,
    height=4cm,
    xtick = {.1,.2,.3,.4,.5,.6,.7,.8,.9},
    xticklabels = {,$\num{.2}$,,$\num{.4}$, ,$\num{.6}$,,$\num{.8}$},
    axis line style={thick},
  ]

    \addplot[KITcyanblue, thick,samples=1000] {floor(min(4 * 5 * abs(x-0.25),4))};
    \addplot[KITorange, thick,samples=1000] {floor(min(4 * 5 * abs(x-0.5),4))};
    \addplot[KITpurple, thick,samples=1000] {floor(min(4 * 5 * abs(x-0.75),4))};

    \addlegendentry{$j=1$}
    \addlegendentry{$j=2$}
    \addlegendentry{$j=3$}
    \addlegendimage{dashed, black}\addlegendentry{$t_j^{(\mathrm{f})}(x)$};
    \addlegendimage{solid, black}\addlegendentry{$k_j^{(\mathrm{f})}(x)$};

    \addplot[KITpurple, dashed, thick] {min(4 * 5 * abs(x-0.75),4)};
    \addplot[KITorange, dashed, thick] {min(4 * 5 * abs(x-0.5),4)};
    \addplot[KITcyanblue, dashed, thick] {min(4 * 5 * abs(x-0.25),4)};

  \end{axis}
\end{tikzpicture}}
    \caption{Continuous- and discrete-time encoding characteristics for $J=3$ \ac{ttfs} encodings with~${T_\mathrm{max}=\Kmax=4}$. %
    }
    \label{fig:encoding}
\end{figure}

The timing ${t_j^{(\mathrm{f})}}$ of the firing of the first spike of the $j$-th signal can be obtained by ${t_j^{(\mathrm{f})}(x)=f_j(x)}$, where ${f_j(\cdot):\mathbb{R}\rightarrow \mathbb{R}}$ is an arbitrary real-valued function. 
In the following, we use ${f_j(x) = \min\left(\alpha_j \left\vert x-\chi_j \right\vert,\, T_\mathrm{max}\right)}$~\cite{arnold_journal}, where $\alpha_j\in\mathbb{R}$ and $\chi_j\in\mathbb{R}$ are the slope and reference of $f_j(\cdot)$ and $T_\mathrm{max}$ the maximum signal duration.
When emulating the encoding and \acp{snn} using a fixed time grid, the spike signals are time-discretized, where $k$ denotes the discrete time index.
Consequently, the timing of the spikes are also discretized, ${k_j^{(\mathrm{f})}(x) =  \big\lfloor \min\left(\alpha_j \left\vert x-\chi_j \right\vert,\, K\right) \big\rfloor}$
where $K$ denotes the maximum number of time steps.
Fig.~\ref{fig:encoding} shows exemplary characteristic curves with $J=3$.
For the discrete-time encoding there exist multiple intervals of width $0.05$, where the encoding output looks alike. Thus, the resolution of the encoded data is limited.

The resolution can be increased by increasing $K$ or $J$, or by finding proper parameters of $f_j(\cdot)$, by, e.g., optimization.
While the former increases the emulation time or the size of the \ac{snn}, the latter assumes that we can optimize the encoding. 
However, while $f_j(\cdot)$ may be a differentiable function, converting $t_j^{(\mathrm{f})}$ into a spike signal $\bm{z}_j(t)$ is not differentiable.
Thus, the backpropagation algorithm cannot be used to update the encoding.

\newpage

\section{Policy-Gradient based Optimization}
\newcommand{\bt}{\bm{\theta}}
To optimize the encoding, the \ac{rl}-based method of~\cite{aoudia18} can be used.
In~\cite{aoudia18}, the parameters of an \ac{ann}-based transmitter prior to a non-differentiable channel are updated using the \ac{pgt}~\cite{PGT_sutton}.
It can be described as a trial-and-error approach: By sampling from a Gaussian distribution, the output of the \ac{ann} is altered multiple times, its performance is obtained, and used to update the parameters of the \ac{ann}.

Let $\bm{\theta} \in \mathbb{R}^N$ denote a parameter vector.
We can now sample $\tilde{\bt}$ from a Gaussian distribution ${\pi(\tilde{\bm{\theta}}|\bm{\theta})=\frac{1}{(\pi \sigma_\pi^2)^N}\exp \left(-\frac{|| \tilde{\bm{\theta}}-\bm{\theta}||_2^2}{\sigma_\pi^2} \right)}$ and measure its performance ${\ell_{\tilde{\bt}}}$ on the given task by using, e.g., the \ac{ce} loss. 
We further define the loss function as ${J(\bt,\pi)=\int_{\mathbb{R}^N} \ell_{\tilde{\bt}} \, \pi(\tilde{\bt}_b|\bt) }\,\mathrm{d}\tilde{\bt}$, which is the expected performance when following $\pi(\tilde{\bt}_b|\bt)$.
We aim to find $\bt$ that minimizes $J(\bt,\pi)$.
Therefore, we estimate the gradient of $J(\bt,\pi)$ by
\vspace*{-2mm}
\begin{align*}
    \nabla_{\bt} J(\bt,\pi) &\stackrel{\text{(a)}}{=}  \mathbb{E}_\pi \left\{ \ell_{\tilde{\bt}} \nabla_{\bt} \ln \left( \pi(\tilde{\bt}|\bt) \right) \right\} \\
    &\stackrel{\text{(b)}}{=} \mathbb{E}_\pi \left\{\ell_{\tilde{\bt}} \frac{2}{N\sigma_\pi^2} \left( \tilde{\bt}-\bt \right) \right\} \\
    &\stackrel{\text{(c)}}{\approx} \frac{1}{B} \sum_{b=1}^B \ell_b \frac{2}{N\sigma_\pi^2} \left( \tilde{\bt}_b-\bt \right) \, , 
\end{align*}
where in (a) we exploit the \ac{pgt}~\cite[pp.~325-326]{sutton18}, in (b) we insert ${\pi(\tilde{\bm{\theta}}|\bm{\theta})}$, and in (c) we use the law of large numbers by sampling $B \in\mathbb{N}$ variations ${\tilde{\bt}_b,\, b=1,\ldots,B}$, with $\ell_b:=\ell_{\tilde{\bt}_b}$.
Using, e.g., \ac{sgd} ${\bt \leftarrow \bt - \epsilon \nabla_{\bt} J(\bt,\pi)}$, where $\epsilon$ denotes the learning rate, $\bt$ can be optimized.

This update rule is pretty intuitive: The policy varies $\bt$ multiple times. 
For each $\tilde{\bt}_b$, its performance $\ell_b$ is obtained, which scales the contribution $\left( \tilde{\bt}_b-\bt \right)$ to the overall update.

To stabilize training, we modify the \ac{sgd} update
\vspace*{-2mm}
\begin{align*}
    \bt \leftarrow  \frac12\left( \bt - \epsilon \sum_{b=1}^B \frac{\ell_b-\ell_{\bt}}{\ell_{\bt}} \left( \tilde{\bt}_b-\bt \right) + \bt^\star\right) \, ,
\end{align*}
where $\ell_{\bt}$ denotes the performance of $\bt$ and $\bt^\star$ the best performing $\bt$ found so far.
We include the constant factor $ \frac{2}{BN\sigma_\pi^2}$ in the learning rate $\epsilon$, and update using $\frac{\ell_b-\ell_{\bt}}{\ell_{\bt}}$ instead of $\ell_b$.
If $\ell_b-\ell_{\bt}<0$, $\tilde{\bt}_b$ outperforms $\bt$, and the \ac{sgd} update will move $\bt$ towards $\tilde{\bt}_b$.
If $\ell_b-\ell_{\bt}>0$, $\tilde{\bt}_b$ performs worse than $\bt$, and the \ac{sgd} update will move $\bt$ away from $\tilde{\bt}_b$.
Furthermore, the denominator normalizes the amplitude of the update.
With increasing or decreasing $\ell_{\bt}$, the update does not vanish or explode.
With the greedy term $\bt^\star$, we were able to avoid oscillations during optimization.  

\newcommand{\Ni}{N_\mathrm{in}}
\newcommand{\Nhid}{N_\mathrm{hid}}
\newcommand{\No}{N_\mathrm{out}}

\vspace*{-2mm}
\section{Parameters of Implementation}
Following~\cite{arnold_journal}, the transmission of 4-\ac{pam} symbols over an \ac{imdd} link is simulated, where the \ac{imdd} link has a baud rate of $112\,\mathrm{GBd}$, a wavelength of $\SI{1270}{\nano\meter}$ and a dispersion coefficient of $-5\,\mathrm{ps\,nm^{-1}\,km^{-1}}$. 
We benchmark our approach against the LCD task of~\cite{nice}, which reproduces~\cite{arnold_journal}. 
While the \ac{snn} is simulated and updated using the PyTorch-based \ac{snn} framework \texttt{norse}~\cite{norse}, the parameters of the neural encoding are updated using \ac{pgt}-based optimization. 
The \ac{snn} consists of $\Ni=n_\mathrm{tap}J$ input neurons, where $n_\mathrm{tap}=7$ is the number of equalizer taps, $\Nhid=40$ hidden layer spiking \ac{lif} neurons, and $\No=4$ non-spiking \ac{li} output neurons.
The linear layers connecting the neurons do not add any bias.
The membrane and synapse time constant of the \ac{lif} and \ac{li} neurons are ${\tau_\mathrm{m}=\tau_\mathrm{s}=\SI{6}{\milli\second}}$, the threshold is ${v_\mathrm{th}=1}$, and the time resolution is ${\Delta t=\SI{0.5}{\milli\second}}$~\cite{arnold_journal}.
All \acp{snn} are optimized using \ac{bptt} with a batch size of $100\,000$, a learning rate of $10^{-3}$, $40\,000$ epochs and using the \ac{ce} loss.
To optimize the encoding, we fix $J$ and $\Kmax$, and initialize $\chi_j=j\frac{y_\mathrm{max}}{J}$ and $\alpha_j=\alpha=\frac{y_\mathrm{max}\Kmax}{6}$, where $y_\mathrm{max}=7$ is the largest value we expect to be output by the \ac{imdd} link. 
During the first $10\,000$ epochs, both the \ac{snn} and the encoding are updated in turns and the encoding is fixed afterward.
For each update of the encoding, $\alpha_j$ and $\chi_j$ are varied $B=20$ times with $\sigma^2_\pi=0.01$, and updated using $\epsilon=0.5$.
As performance measure $\ell_b$, the \ac{ce} loss is used.

At a channel noise power of $\sigma_\mathrm{n}^2=\SI{-20}{\decibel}$, we optimized various encodings for ${J=4,6,8,10}$ and ${\Kmax=4,6,8,10}$.
We compare their \ac{ber}, average number $Z_\mathrm{avg}$ of spikes generated per inference, and the average number of \ac{mac} operations per inference when emulated on digital hardware.
The number of \ac{mac} operations is given by ${\#\mathrm{\ac{mac}}=\Nhid(7J+4)\cdot\Kmax}$~\cite{moursi_fpga}, where ${\Nhid(7J+4)}$ is the number of parameters of the linear layers and thus the number of adaptable \ac{snn}-parameters.

As a benchmark, we plot the \ac{ber} obtained from~\cite{arnold_journal,nice} with $J=10,\, \Kmax=60$, in the following denoted ``$\mathrm{ref}$''. 
In addition, for various $J$ we optimized \acp{snn} combined with the non-optimized encoding of~\cite{arnold_journal} ($\Kmax=60$), which in the following we refer to as~${K=60,\, \mathrm{NoOpt}}$. 
The architectures of $\mathrm{ref}$ and ${K=60,\, \mathrm{NoOpt}}$ with $J=10$ are identical; the only difference lies in their training configurations. Specifically, $\mathrm{ref}$ was trained using a batch size of 100 samples per update, whereas ${K=60,\, \mathrm{NoOpt}}$ was trained with a batch size of~$100\,000$. 

\vspace*{-2mm}
\section{Results}
\begin{figure*}
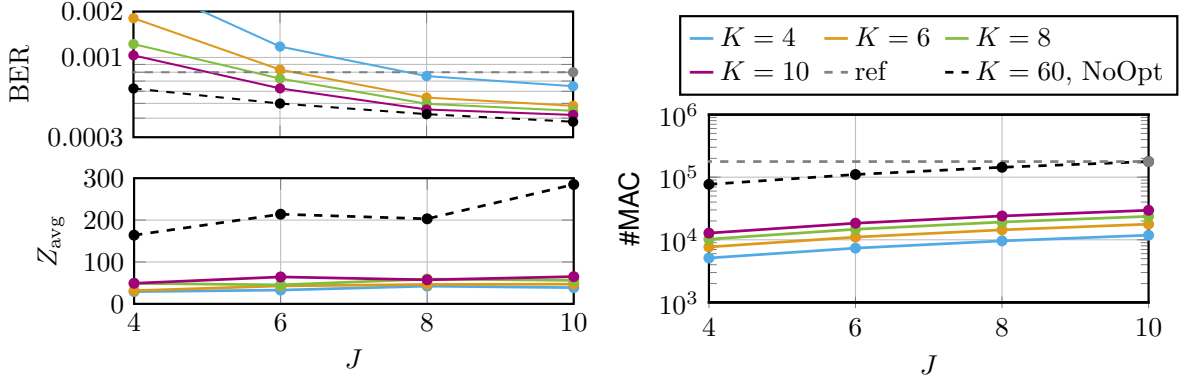

    \centering
    \resizebox{.99\textwidth}{!}{\begin{tikzpicture}[thick]

    \def\lwidth{.75}
    \def\opac{100}
    \def\opaci{100}
    \def\marksz{1.5pt}

    \node at (0, 2.5) {\input{figures/ber_arnold.tex}};
    
    \node at (0.27, 0) {\input{figures/spike_rate.tex}};
    
    \node at (7.7, .9) {\input{figures/mac.tex}};           %

    \end{tikzpicture}}
    \vspace*{-2mm}
    \caption{Comparison of the optimized encodings for various number of discrete time steps $K$ and number of spike signals $J$, with the reference $\mathrm{ref}$ of~\cite{arnold_journal,nice}, and ${K=60,\,\mathrm{NoOpt}}$, which is an optimization of~\cite{arnold_journal} without minimization of $Z_\mathrm{avg}$.
    All approaches are compared regarding their \ac{ber}, average number of generated spikes $Z_\mathrm{avg}$ and number of \ac{mac} operations $(\# \mathrm{MAC})$.}
    \label{fig:results}
\end{figure*}
Fig.~\ref{fig:results} shows the results, where each dot denotes a simulated data point, evaluated using $10^7$ samples.
Comparing $\mathrm{ref}$ and ${K=60,\, \mathrm{NoOpt}}$ with $J=10$, we observed that increasing the batchsize of the training greatly improves performance.
While the non-optimized encoding $(K=60,\, \mathrm{NoOpt})$ achieves the best performance, it comes at the cost of a large $Z_\mathrm{avg}$.
If the encoding is optimized, all tested combinations of $J$ and $K$ achieve a similar $Z_\mathrm{avg}$, significantly lower than $Z_\mathrm{avg}$ of $(K=60,\, \mathrm{NoOpt})$.
Depending on the combination of $J$ and $K$, this is accompanied by a small degradation in performance.
Due to the massive reduction of $K$, the runtime of the \ac{snn} emulation and thus \#\ac{mac} is drastically reduced.

For $K=10,\, J=10$, we achieve the lowest \ac{ber}. 
Compared to ${K=60,\, \mathrm{NoOpt}}$ with $J=10$, the $K=10,\, J=10$ setup reduces $Z_\mathrm{avg}$ by $77.2\%$, and both $\Kmax$ and \#\ac{mac} by $83.3\%$, with a \ac{ber} penalty of only $\num{4.1}\cdot10^{-5}$.
The number of adaptable \ac{snn}-parameters is alike.
Decreasing $K$ and $J$, we can further reduce $Z_\mathrm{avg}$, \#\ac{mac} and the number of \ac{snn}-parameters.
The $K=6,\, J=8$ setup reduces $Z_\mathrm{avg}$ by $83.8\%$, $\Kmax$ by $90\%$, the number of adaptable \ac{snn}-parameters by $18.9\%$, and \#\ac{mac} by $91.9\%$, with a \ac{ber} penalty of~${\num{1.6}\cdot10^{-4}}$.

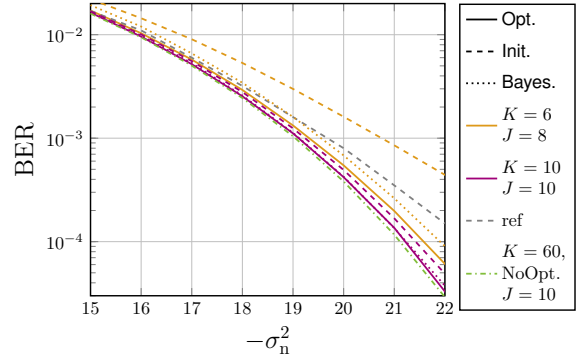
\begin{figure}
    \centering
    \resizebox{.47\textwidth}{!}{\begin{tikzpicture}[thick,>=latex]
    \def\lwidth{1.0}
    \def\opac{100}
    \def\opaci{100}
    \def\marksz{1.5pt}

    \begin{axis}[
        thick,
        ymode = log,
        xscale=1,
        xlabel=\Large $-\sigma_\mathrm{n}^2$,
        ylabel={\Large $\mathrm{BER}$},
        xtick = {15,16,17,18,19,20,21,22,23},
        grid=major,
        legend cell align={left},
        legend style={
            at={(1.04,1)},
            anchor=north west,
            fill opacity = 0.8,
            draw opacity = 1, 
            text opacity = 1,
            align=left,
            row sep=5pt
        },
        xmin=15,xmax=22,
        ymin=3e-5,ymax=2e-2,
        axis line style=thick,
        tick label style={/pgf/number format/fixed},
        ]
    
    \addlegendimage{solid, black,line width=\lwidth}\addlegendentry{Opt.};
    \addlegendimage{dashed,black,line width=\lwidth}\addlegendentry{Init.};
    \addlegendimage{dotted,black, line width=\lwidth}\addlegendentry{Bayes.};

    \addplot[color=KITorange,line width=\lwidth] table [x=var,y = 6_8] {figures/txt/ber.txt};
    \addlegendentry{$K=6$\\$J=8$};

    \addplot[color=KITpurple,line width=\lwidth] table [x=var,y = 10_10] {figures/txt/ber.txt};
    \addlegendentry{$K=10$\\$J=10$};

    \addplot[color=gray, dashed, line width=\lwidth] coordinates{(15,1.7e-2) (16,1.1e-2) (17,6e-3) (18,3.2e-3) (19,1.6e-3) (20,8e-4) (21,3.5e-4) (22,1.5e-4)};
    \addlegendentry{$\mathrm{ref}$}

    \addplot[color=KITpalegreen, dashdotted, line width=\lwidth] table [x=var, y=noopt_60_60] {figures/txt/ber.txt};
    \addlegendentry{$K=60,$\\$\mathrm{NoOpt.}$\\ $ J=10$};

    \addplot[color=KITorange,dashed,line width=\lwidth] table [x=var,y = noopt_6_8] {figures/txt/ber.txt};

    \addplot[color=KITorange, dotted, line width=\lwidth] coordinates {(23,3.06e-5) (22,9.04e-5) (21,2.624e-4) (20,6.811e-4) (19,1.606e-3) (18,3.4342e-3) (17,6.5965e-3) (16,1.2e-2) (15,1.9291e-2)};

    \addplot[color=KITpurple,dashed,line width=\lwidth] table [x=var,y = noopt_10_10] {figures/txt/ber.txt};

    \addplot[color=KITpurple, dotted, line width=\lwidth] coordinates {(23,8.05e-06) (22,3.75e-05) (21,1.3523e-04) (20,4.2355e-04) (19,1.1262e-03) (18,2.5841e-03) (17,5.3159e-03) (16,9.811e-03) (15,1.6652e-02)};

    \end{axis}
\end{tikzpicture}}
    \caption{\ac{ber} of \acp{snn} with encodings optimized using the proposed approach (Opt.), using Bayesian optimization (Bayes.), and initialized encodings without optimization (Init.) for two combinations of $J$ and $K$. As reference $\mathrm{ref}$, \cite{arnold_journal} and ${K=60,\, \mathrm{NoOpt}}$ with $J=10$ are displayed.}
    \label{fig:ber}
\end{figure}

Fig.~\ref{fig:ber} shows the \ac{ber} of optimized \acp{snn} when combined with encoders with initial and optimized parameters.
We further obtained the parameters through Bayesian optimization, based on 100 \acp{snn} evaluations.
The Bayesian optimization was initialized in the same way as the proposed approach and was provided by~\cite{wandb}.
While for $K=10,\,J=10$ the initial parameters already perform well, for a reduced number of parameters $K=6,\,J=8$, our approach is able to significantly improve the encoding. While Bayesian optimization also improves the initial parameters, it is outperformed by our approach.  
It is important to note that Bayesian optimization using 100 \acp{snn} took fifty times longer than our proposed approach.
For $K=60,\, J=10$ we also optimized an encoding, which performs alike as ${K=60,\, \mathrm{NoOpt}}$ with $J=10$.
This suggests, that the encoding parameters of~\cite{arnold_journal} were well chosen.

We conclude that the proposed approach exhibits multiple benefits:
First, time-consuming heuristic optimization of the neural encoding parameters can be avoided. 
Second, an efficient neural encoding is learned, reducing both the runtime and size of the \ac{snn}.
Third, compared to~\cite{arnold_journal}, for a constant $Z_\mathrm{avg}$, the performance of the overall system is improved.
Fourth, to optimize the parameters of the encoding, no gradient needs to be backpropagated through the \ac{snn} and encoding. 

\vspace*{-2mm}
\section{Conclusion}
In this paper, we proposed an \ac{rl}-based method to optimize the parameters of the neural encoding.
For an \ac{imdd} link, neural encoding, and \ac{snn} of~\cite{arnold_journal}, we showed that the proposed approach can reduce the computational load, the number of parameters of the \ac{snn}-based equalizer and the number of generated spikes, with only a minor \ac{ber} penalty.

\clearpage
\section{Acknowledgments}
This work has received funding from the European Research Council (ERC) under the European Union’s Horizon 2020 research and
innovation programme (grant agreement No. 101001899).
\defbibnote{myprenote}{%
}

\printbibliography[prenote=myprenote]

\vspace{-4mm}

\end{document}